\documentclass[10pt,letterpaper]{article}

\usepackage[hidelinks]{hyperref}

\usepackage{cogsci}

\cogscifinalcopy 

\usepackage{pslatex}
\usepackage{apacite}
\usepackage{float} 

\usepackage{covington}
\usepackage{graphicx}

\title{Can Peanuts Fall in Love with Distributional Semantics?}
 
\author{{\large \bf James A. Michaelov (j1michae@ucsd.edu)}\\{\large \bf Seana Coulson (scoulson@ucsd.edu)}\\{\large \bf Benjamin K. Bergen (bkbergen@ucsd.edu)} \\
Department of Cognitive Science, University of California, San Diego\\
9500 Gilman Dr, La Jolla, CA 92093, USA}

\begin{document}

\maketitle

\begin{abstract}
Context changes expectations about upcoming words---following a story involving an anthropomorphic peanut, comprehenders expect the sentence \textit{the peanut was in love} more than \textit{the peanut was salted}, as indexed by N400 amplitude \cite{nieuwland_2006_WhenPeanutsFall}. This updating of expectations has been explained using Situation Models---mental representations of a described event. However, recent work showing that N400 amplitude is predictable from distributional information alone raises the question whether situation models are necessary for these contextual effects. We model the results of \citeA{nieuwland_2006_WhenPeanutsFall} using six computational language models and three sets of word vectors, none of which have explicit situation models or semantic grounding. We find that a subset of these can fully model the effect found by \citeA{nieuwland_2006_WhenPeanutsFall}. Thus, at least some processing effects normally explained through situation models may not in fact require explicit situation models.

\end{abstract}

\section{Introduction}

It is widely believed that prediction plays a key role in language processing, with more predictable words being processed more easily \cite{fischler_1979_AutomaticAttentionalProcesses,kutas_1984_BrainPotentialsReading,levy_2008_ExpectationbasedSyntacticComprehension,kutas_2011_LookWhatLies,vanpetten_2012_PredictionLanguageComprehension,delong_2014_PreProcessingSentenceComprehension,luke_2016_LimitsLexicalPrediction,kuperberg_2020_TaleTwoPositivities}. Perhaps the strongest evidence for this comes from the N400, a neural signal of processing difficulty that is highly correlated with lexical probability---contextually probable words elicit an N400 response of smaller (less negative) amplitude than contextually improbable words, whether predictability is determined based on human judgements (\citeNP{kutas_1984_BrainPotentialsReading}; for review see \citeNP{vanpetten_2012_PredictionLanguageComprehension}) or a corpus \cite{parviz_2011_UsingLanguageModels,frank_2015_ERPResponseAmount,aurnhammer_2019_EvaluatingInformationtheoreticMeasures,merkx_2021_HumanSentenceProcessing,szewczyk_2022_ContextbasedFacilitationSemantic,michaelov_2022_ClozeFarN400,michaelov_2023_StrongPredictionLanguage}.

A striking feature of the predictions indexed by the N400 is how flexible they can be. Under normal circumstances, a sentence such as \textit{the peanut was in love} would be highly improbable, much more so than \textit{the peanut was salted}. Following the short story in (\ref{ex:full_stim}), however, this changes \cite{nieuwland_2006_WhenPeanutsFall}.

\begin{example}
\label{ex:full_stim} 
A woman saw a dancing peanut who had a big smile on his face. The peanut was singing about a girl he had just met. And judging from the song, the peanut was totally crazy about her. The woman thought it was really cute to see the peanut singing and dancing like that.
\end{example}

In fact, \citeA{nieuwland_2006_WhenPeanutsFall}, who tested this in Dutch, found that in the context of (\ref{ex:full_stim}), the last word of \textit{de pinda was \textbf{verliefd}} (`the peanut was \textbf{in love}') elicited a smaller N400 than \textit{de pinda was \textbf{gezouten}} (`the peanut was \textbf{salted}'). How does such a dramatic reversal occur?

One possibility put forward by \citeA{nieuwland_2006_WhenPeanutsFall} is that while reading the context, the reader's mental representation of the peanut is altered such that it is treated as an animate entity. This, as \citeA{nieuwland_2006_WhenPeanutsFall} note, is in line with theories of situation models, which argue that we track the entities under discussion, as well as their properties and relations. Such accounts generally involve explicit structures or schemata, grounding in world knowledge or experience, extraction of propositional information, or a combination of these (see, e.g., \citeNP{bransford_1972_SentenceMemoryConstructive,kintsch_1978_ModelTextComprehension,johnson-laird_1980_MentalModelsCognitive,garnham_1981_MentalModelsRepresentations,johnson-laird_1983_MentalModelsCognitive,vandijk_1983_StrategiesDiscourseComprehension,kintsch_1988_RoleKnowledgeDiscourse,zwaan_1995_ConstructionSituationModels,zwaan_1995_DimensionsSituationModel,radvansky_1998_RetrievalTemporallyOrganized,kintsch_1998_ComprehensionParadigmCognition,zwaan_1998_SituationModelsLanguage,zwaan_2004_UpdatingSituationModels,kintsch_2005_OverviewTopDownBottomUp,vanberkum_2007_EstablishingReferenceLanguage,kintsch_2011_ConstructionMeaning,butcher_2012_TextComprehensionDiscourse,zwaan_2014_EmbodimentLanguageComprehension,zwaan_2016_SituationModelsMental,zacks_2016_DiscourseComprehension,kintsch_2018_RevisitingConstructionIntegration,hoebenmannaert_2021_SituationModelUpdating}). On a situation model account, the reader alters their semantic representation of the peanut to give it animate features in accordance with the information that it can sing, dance, and show emotions, thereby facilitating the processing of \textit{in love}.

The hypothesis that structured or grounded situation models explain N400 effects such as those found by \citeA{nieuwland_2006_WhenPeanutsFall} is generally accepted (e.g.,  \citeNP{hagoort_2007_SentenceGiven,filik_2008_ProcessingLocalPragmatic,warren_2008_EffectsContextEye,rosenbach_2008_AnimacyGrammaticalVariation,ferguson_2008_AnomaliesRealCounterfactual,ferguson_2008_EyemovementsERPsReveal,menenti_2009_WhenElephantsFly,bicknell_2010_EffectsEventKnowledge,degroot_2011_LanguageCognitionBilinguals,metusalem_2012_GeneralizedEventKnowledge,aravena_2014_ActionRelevanceLinguistic,zwaan_2014_EmbodimentLanguageComprehension,xiang_2015_ReversingExpectationsDiscourse,kuperberg_2020_TaleTwoPositivities}) and has been shown to be viable using computational models \cite{venhuizen_2019_ExpectationbasedComprehensionModeling}. However, there are alternative explanations. 

The present study asks whether the effect can instead be explained by lexical preactivation based on distributional linguistic knowledge, following the findings that the statistics of language can be used to model N400 effects \cite{ettinger_2016_ModelingN400Amplitude,michaelov_2020_HowWellDoes,michaelov_2021_DifferentKindsCognitive,michaelov_2022_CollateralFacilitationHumans,uchida_2021_ModelOnlineTemporalSpatial,michaelov_2023_StrongPredictionLanguage} and predict single-trial N400 amplitude \cite{chwilla_2005_AccessingWorldKnowledge,parviz_2011_UsingLanguageModels,vanpetten_2014_ExaminingN400Semantic,frank_2015_ERPResponseAmount,aurnhammer_2019_ComparingGatedSimple,aurnhammer_2019_EvaluatingInformationtheoreticMeasures,merkx_2021_HumanSentenceProcessing,michaelov_2021_DifferentKindsCognitive,szewczyk_2022_ContextbasedFacilitationSemantic,michaelov_2023_StrongPredictionLanguage}.

Specifically, we look at two possible ways in which this might arise. One, which we refer to as \textit{event-level priming}, refers to the idea that a word associated with a previously-discussed event may be more likely to be predicted by virtue of this. This is something that has been previously reported in the N400---\citeA{metusalem_2012_GeneralizedEventKnowledge}, for example, found that that merely being related to the event under discussion leads to a smaller N400 response to a word even when that word is inappropriate. \citeA{michaelov_2022_CollateralFacilitationHumans} model this with transformer language models---systems trained to calculate the probability of a word given its context based on the statistics of language alone---showing that this effect is explainable with distributional information. Thus, it may be the case that the fact that \textit{in love} is related to, for example, being \textit{crazy about} someone that leads to it being predicted to be more likely than \textit{salted}. Following \citeA{michaelov_2022_CollateralFacilitationHumans}, we investigate this using 6 Dutch transformer language models \cite{havinga_2021_GPT2MediumPretrainedCleaned,havinga_2022_GPT2LargePretrainedCleaned,havinga_2022_GPTNeo125MPretrained,havinga_2022_GPTNeo3B,devries_2019_BERTjeDutchBERT,delobelle_2020_RobBERTDutchRoBERTabased}, testing whether they show the same effect as humans---that is, whether they predict the canonical sentence the \textit{peanut was salted} to be less likely than the noncanonical sentence \textit{the peanut was in love}.

An alternative possibility is \textit{lexical priming}. More simply than in the case of event-level priming, it may be the case that the preceding context involving words such as \textit{dancing}, \textit{smile}, \textit{singing}, \textit{crazy}, and \textit{cute} exerts a stronger pressure on prediction of \textit{in love} than \textit{peanut} does on \textit{salted}. Intuitively, one might expect that a system (neurocognitive or computational) displaying event-level priming is likely to display lexical priming---indeed, lexical priming is a possible mechanism by which at least some part of event-level priming could be achieved. The fact that lexical priming is likely to occur in a system displaying event-level priming is also supported by the fact that language models show both \cite{kassner_2020_NegatedMisprimedProbes,misra_2020_ExploringBERTSensitivity,michaelov_2022_CollateralFacilitationHumans}. Thus, in the present study, we distinguish between two possible explanations of the effect found by \citeA{nieuwland_2006_WhenPeanutsFall}: lexical priming alone, and event-level priming that may include lexical priming. 

As discussed, language models can be used to model the latter. To model the former, we turn to word vectors---representations of words derived from their co-occurrence statistics, either directly or based on word embeddings learned by neural networks (see, e.g., \citeNP{dumais_1988_UsingLatentSemantic,landauer_1998_IntroductionLatentSemantic,mikolov_2013_DistributedRepresentationsWords,pennington_2014_GloveGlobalVectors,mikolov_2018_AdvancesPreTrainingDistributed,tulkens_2016_EvaluatingUnsupervisedDutch,grave_2018_LearningWordVectors}). The cosine distance between the vector of each critical word (e.g. \textit{in love} or \textit{salted}) and the mean of the vectors of the words in the preceding context can therefore be used to test how similar the critical word is to the words preceding it \cite{ettinger_2016_ModelingN400Amplitude,uchida_2021_ModelOnlineTemporalSpatial}, and thereby model the effects of lexical priming alone. To do this this we used three sets of Dutch word vectors (from \citeNP{tulkens_2016_EvaluatingUnsupervisedDutch}; and \citeNP{grave_2018_LearningWordVectors}).

\section{Background} 
A number of researchers have attempted to model the N400 computationally, including using language models \cite{parviz_2011_UsingLanguageModels,frank_2015_ERPResponseAmount,aurnhammer_2019_EvaluatingInformationtheoreticMeasures,michaelov_2020_HowWellDoes,merkx_2021_HumanSentenceProcessing,michaelov_2021_DifferentKindsCognitive,michaelov_2022_ClozeFarN400,szewczyk_2022_ContextbasedFacilitationSemantic,michaelov_2023_StrongPredictionLanguage} and the distances between vector representations of words \cite{parviz_2011_UsingLanguageModels,vanpetten_2014_ExaminingN400Semantic,ettinger_2016_ModelingN400Amplitude,uchida_2021_ModelOnlineTemporalSpatial,michaelov_2023_StrongPredictionLanguage}. There have also been several attempts to computationally model whether the amplitude of the N400 response is impacted by situation models \cite{uchida_2021_ModelOnlineTemporalSpatial,venhuizen_2019_ExpectationbasedComprehensionModeling} and thematic roles \cite{brouwer_2017_NeurocomputationalModelN400,fitz_2019_LanguageERPsReflect,rabovsky_2018_ModellingN400Brain}.

To our knowledge, only one previous study \cite{uchida_2021_ModelOnlineTemporalSpatial} has directly attempted to model the discourse effect found by \citeA{nieuwland_2006_WhenPeanutsFall}, and it does not rely on purely distributional linguistic information. \citeA{uchida_2021_ModelOnlineTemporalSpatial} base their model on Wikipedia2Vec \cite{yamada_2020_Wikipedia2VecEfficientToolkit} vectors---while these include distributional information derived from the surface-level statistics of language, they also include information about hyperlinks between Wikipedia pages, and thus structured semantic relations based on human judgements of relevance and importance \cite{yamada_2020_Wikipedia2VecEfficientToolkit}. Additionally, \citeA{uchida_2021_ModelOnlineTemporalSpatial} only look at the English-translated version of the single stimulus item presented in (1), and thus, it is unclear whether the results generalize to all the stimuli in the original study. The current study overcomes these inferential limitations by using the original Dutch stimuli and by using neural language models and word vectors trained only on natural language input.

\section{The present study}

We investigate the adequacy of distributional knowledge to explain the human N400 effect found by \citeA{nieuwland_2006_WhenPeanutsFall} using predictions of neural network language models and the distance between the word vectors of the critical words and their context. Specifically, we ask this question for two possible variants of the effect found by \citeA{nieuwland_2006_WhenPeanutsFall}.

\citeA{nieuwland_2006_WhenPeanutsFall} presented experimental participants with short stories such as those in (\ref{ex:full_stim}) including ``canonical'' sentences like  \textit{the peanut was \textbf{salted}} or ``noncanonical'' ones like \textit{the peanut was \textbf{in love}}. One approach to whether language models and humans show the same prediction patterns (taken by \citeNP{uchida_2021_ModelOnlineTemporalSpatial}) is to compare the statistical metrics the critical words elicit in the context of the full story versus in isolation. Without  preceding context, these sentences should produce values that match the canonicality of the sentence, but the difference should attenuate or reverse following the story context. 

Thus, we ran a statistical analysis testing for an interaction between stimulus length (full story or only the last sentence) and canonicality (canonical or noncanonical). Such an interaction would reveal a context-dependent difference in the effect of canonicality on our  statistical metrics; and thus would replicate in neural language models the effect found by \citeA{nieuwland_2006_WhenPeanutsFall}. 

However, an interaction between stimulus length and canonicality in this direction could result from either a reversal or a decrease in the magnitude of the canonicality effect. Canonical stimuli might elicit lower surprisals or smaller cosine distances in both context conditions, but of different magnitudes. For this reason, we label the effect measured by an interaction (in the expected direction) a \textbf{reduction effect}.

\citeA{nieuwland_2006_WhenPeanutsFall} did not employ the 2 x 2 design that would allow them to detect an interaction---they compared the N400 in context only, finding that canonical stimuli actually elicited larger N400 responses than noncanonical stimuli. To replicate this finding, we test whether the canonical full-length stimuli elicit higher surprisals or greater cosine distances than the noncanonical full-length stimuli, a \textbf{reversal effect}.

If either language models or word vectors can successfully model the reversal effect, this would suggest that distributional information is sufficient to explain the data reported by \citeA{nieuwland_2006_WhenPeanutsFall}. Thus, while situation models and extralinguistic  information may be involved in the neurocognitive system underlying the N400, additional evidence is required to prove this. If neither can model either effect, this would undermine the claim that distributional information is sufficient to explain the effect found by \citeA{nieuwland_2006_WhenPeanutsFall}. Finally, if either language models or word vectors can successfully model the reduction effect but not the full reversal effect, this may support the idea that distributional information could be used as part of the neurocognitive system underlying the N400 response, but that it is not sufficient to yield the dynamic contextual sensitivity humans display. Situation models and other sources of information might explain the remainder.

\section{Method}
\subsection{Materials}
Stimuli were used from the original experiment, and are provided online\footnote{\url{https://www.researchgate.net/publication/268208198}} by the authors \cite{nieuwland_2006_WhenPeanutsFall}.  We compared the effect of experimental
condition on the N400 and on neural network surprisal (as in \citeNP{michaelov_2020_HowWellDoes}) and the cosine similarity between the word vector of the critical word and the mean of the word vectors in its context (as in \citeNP{ettinger_2016_ModelingN400Amplitude}).

The stimuli use 60 full-length story frames, each of which has either a canonical or noncanonical predicate, for 120 unique stories. As the aim is to model human online comprehension processes, the models only used the text before the critical words (e.g., \textit{in love} or \textit{salted}) to predict the critical words, so stories were truncated after the critical word. For the critical sentence stimuli, we isolated the last sentence of these truncated stories, 
including and up to the critical word in each story (e.g., \textit{The peanut was in love}). This produced 240 stimuli, as shown in Table \ref{tab:conditions}.

\begin{table}[h!]
\centering
\renewcommand{\arraystretch}{1.2}
\begin{tabular}{llr}
\hline
\textbf{Predicate Type} & \textbf{Stimulus Length} & \textbf{Count} \\
\hline
Canonical & Full-length & 60\\
Canonical & critical sentence & 60\\
Noncanonical & Full-length & 60\\
Noncanonical & critical sentence & 60\\
\hline 
\end{tabular}
\caption{Experimental stimuli derived from \citeA{nieuwland_2006_WhenPeanutsFall}.}
\label{tab:conditions}
\end{table}

\subsection{Statistical Analysis}
Statistical analysis and data manipulation were carried out in \textit{R} \cite{rcoreteam_2020_LanguageEnvironmentStatistical} using \textit{Rstudio} \cite{rstudioteam_2020_RStudioIntegratedDevelopment} and the \textit{tidyverse} \cite{wickham_2019_WelcomeTidyverse} and \textit{lme4} \cite{bates_2015_FittingLinearMixedeffects} packages. Code, data, and statistical analyses are provided at {\url{https://osf.io/wnj76}}.

\section{Experiment 1: Language Models}
\subsection{Language models}
We used six pretrained models available through the \textit{transformers} package \cite{wolf_2020_TransformersStateoftheArtNatural}. These were all of the available monolingual Dutch language models using standard architectures and training procedures at the time of analysis. Four of these models---Dutch versions of the medium \cite{havinga_2021_GPT2MediumPretrainedCleaned} and large \cite{havinga_2022_GPT2LargePretrainedCleaned} GPT-2 models \cite{radford_2019_LanguageModelsAre} and Dutch versions of the 125 million parameter \cite{havinga_2022_GPTNeo125MPretrained} and 1.3 billion parameter \cite{havinga_2022_GPTNeo3B} GPT-Neo models \cite{black_2021_GPTNeoLargeScale}---were autoregressive, meaning that they are trained to predict a word based only on its preceding context. The remaining two models---BERTje \cite{devries_2019_BERTjeDutchBERT} and RobBERT v2 \cite{delobelle_2020_RobBERTDutchRoBERTabased}, based on BERT  \cite{devlin_2019_BERTPretrainingDeep} and RoBERTa \cite{liu_2019_RoBERTaRobustlyOptimized}, respectively---are masked language models, meaning that they are also trained to predict a word based on the text following the critical word. However, as stated, in the present study, all models were only provided with the context preceding the critical words. We ran the stimuli through each language model, calculating the surprisal of each critical word that was in the model's vocabulary (we restricted our analyses to these items). To do this, we calculated the negative of the logarithm of the probabilities provided for each critical word by each of the language models. We then tested for the reduction and reversal effects with these surprisal values. The language models were run in \textit{Python} \cite{vanrossum_2009_PythonReferenceManual}, using the \textit{PyTorch} \cite{paszke_2019_PyTorchImperativeStyle} implementation of each model, as provided by the \textit{transformers} package \cite{wolf_2020_TransformersStateoftheArtNatural}.

\begin{figure*}[h]
    \centering
    \includegraphics[width=0.9\textwidth]{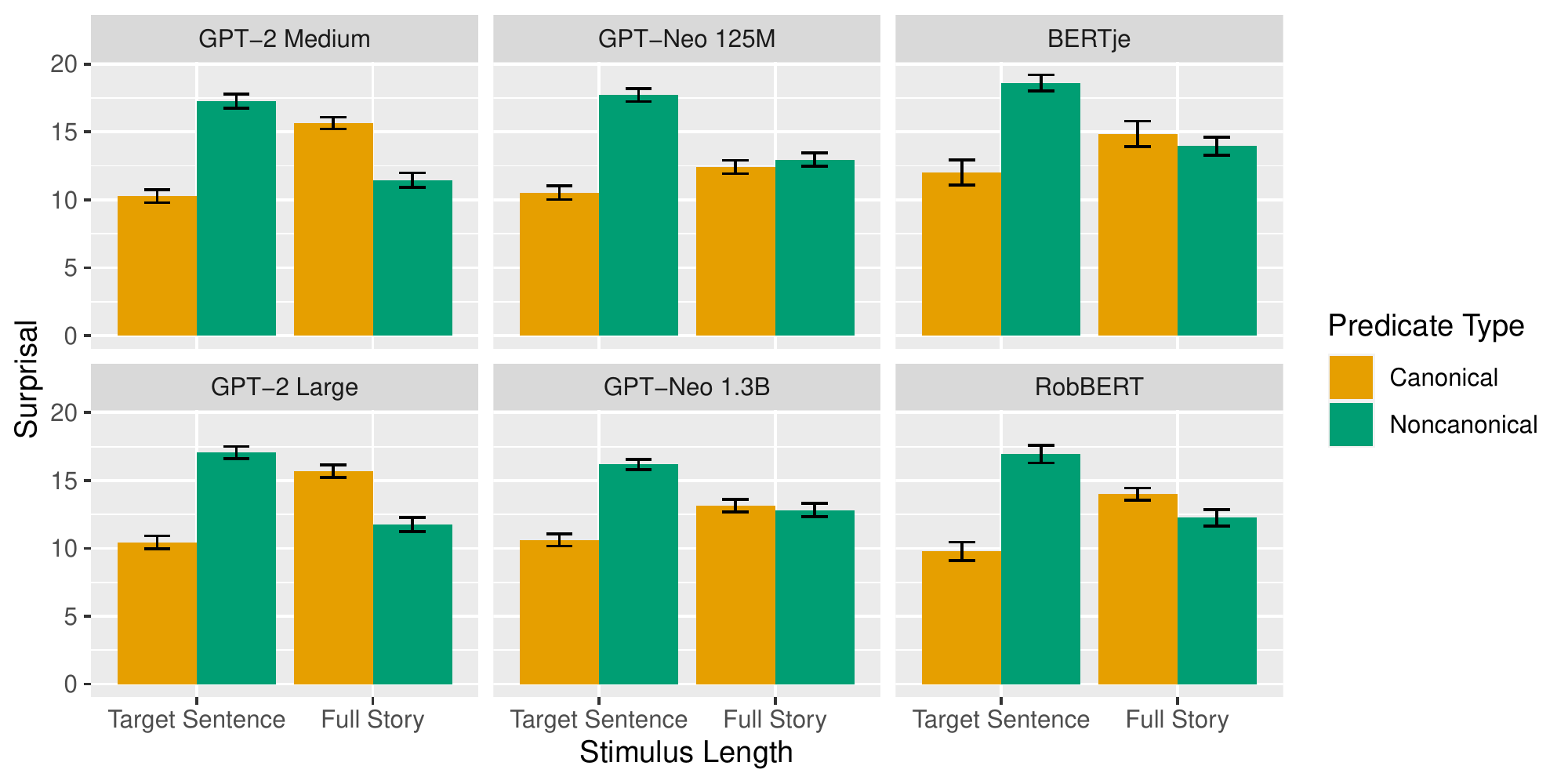}
    \caption{Surprisal elicited by critical words for each predicate type and stimulus length.}
    \label{fig:lms}
\end{figure*}

\subsection{Reduction effect}
In order to test the reduction effect, we constructed linear mixed-effects regression models, with the surprisal calculated from each language model as the dependent variable. In each model, predicate type (canonical or noncanonical) and stimulus length (full-length or critical sentence) were fixed effects and story frame (each of the 60) was a random intercept. For the regressions with the autoregressive models and BERTje surprisal as their dependent variables, we then constructed regressions also including an interaction between predicate type and stimulus length. Using likelihood ratio tests, we found that these regressions including the interaction fit the data significantly better than those without the interaction (GPT-2 Medium: $\chi^2(1) = 112.0, p<0.001$; GPT-2 Large: $\chi^2(1) = 115.9, p<0.001$; GPT-Neo 125M: $\chi^2(1) = 67.3, p<0.001$; GPT-Neo 1.3B: $\chi^2(1) = 56.3, p<0.001$; BERTje: $\chi^2(1) = 44.4, p<0.001$), indicating a significant interaction between predicate type and stimulus length. The regression with RobBERT surprisal as its dependent variable and no interaction had a singular fit, but the regression with the interaction did not. Thus, instead of running a likelihood ratio test to investigate whether there was a significant interaction, we used a a Type III
ANOVA with Satterthwaite's method for estimating degrees of freedom  \cite{kuznetsova_2017_LmerTestPackageTests} on the regression with the interaction, finding it to be a significant predictor of RobBERT surprisal ($F(1,71.2) = 81.1, p<0.001$). Note that all reported $p$-values are corrected for multiple comparisons based on false discovery rate \cite{benjamini_2001_ControlFalseDiscovery}.

For all language models, there was a significant interaction between predicate type and stimulus length. Further inspection of the regressions showed that in all cases, the interaction was in the expected direction. Thus, all models displayed the reduction effect. This can be seen visually in  Figure \ref{fig:lms}---in all models, when only the critical sentence was presented, the mean surprisal for critical words in canonical sentences is lower than for critical words in noncanonical sentences. Conversely, when the full-length story is presented to the language models, the critical words in the noncanonical sentences elicit a lower or roughly-equal surprisal than the critical words in the canonical sentences. 

\subsection{Reversal effect}
To test for which models this latter finding was statistically significant, we initially attempted to fit linear mixed-effects regression models for each the full-length and critical sentence stimulus results for each language model; however, this led to several models with singular fits. Instead, we carried out pairwise two-tailed $t$-tests, comparing the surprisal of canonical and noncanonical stimuli for full-length and critical sentence stimuli for each language model.

First, we test whether the decontextualized canonical critical sentence stimuli elicit significantly lower surprisals than noncanonical critical sentence stimuli. After correction for multiple comparisons, they do so in all language models ({GPT-2 Medium}: $t(88.7)={-9.91}$, $p<0.001$;
{GPT-2 Large}: $t(88.1)={-10.1}$, $p<0.001$;
{GPT-Neo 125M}: $t(88.6)={-10.3}$, $p<0.001$ ;
{GPT-Neo 1.3B}: $t(85.5)={-9.62}$, $p<0.001$;
{BERTje}: $t(48.4)={-5.99}$, $p<0.001$;
{RobBERT}: $t(55.1)={-7.67}$, $p<0.001$).

Next, in order to investigate the reversal effect, we test whether canonical full-length stimuli elicit lower surprisals than noncanonical full-length stimuli.  After correction for multiple comparisons, only the Dutch GPT-2 models successfully model the reversal effect---they are the only models for which canonical full-length stimuli elicit significantly higher surprisals than noncanonical full-length stimuli
({GPT-2 Medium}: $t(86.3)={6.11}$, $p<0.001$;
{GPT-2 Large}: $t(88.4)={5.65}$, $p<0.001$).

The difference in other models was not significant after correction for multiple comparisons 
({GPT-Neo 125M}: $t(88.9)={-0.77}$, $p=1.000$ ;
{GPT-Neo 1.3B}: $t(88.9)={0.47}$, $p=1.000$;
{BERTje}: $t(51.5)={0.79}$, $p=1.000$;
{RobBERT}: $t(46.6)={2.32}$, $p=0.120$).

However, it is worth noting that the contrast between the two sets of results (critical sentence only vs. full stimulus) means that significant canonicality effects for the critical sentence stimuli disappear in the full-length stimuli, underscoring the presence of a reduction effect in the Dutch GPT-Neo models, BERTje, and RobBERT.

\subsection{Discussion}

\citeA{nieuwland_2006_WhenPeanutsFall} found that in a suitably supportive context, noncanonical stimuli like \textit{de pinda was \textbf{verliefd}} (`the peanut was \textbf{in love}') elicit smaller N400 responses than canonical stimuli such as \textit{de pinda was \textbf{gezouten}} (`the peanut was \textbf{salted}')---context not only mitigated but reversed the effect of animacy violation.

We find that two language models also display this reversal effect: Dutch GPT-2 Medium \cite{havinga_2021_GPT2MediumPretrainedCleaned} and Dutch GPT-2 Large \cite{havinga_2022_GPT2LargePretrainedCleaned}. When these models are presented with the same contexts, the surprisal of critical words in the noncanonical condition is lower than that elicited by those in the canonical condition.

This is not the case for the remaining four language models: Dutch GPT-Neo 125M \cite{havinga_2022_GPTNeo125MPretrained}, Dutch GPT-Neo 1.3B \cite{havinga_2022_GPTNeo3B}, BERTje \cite{devries_2019_BERTjeDutchBERT}, and RobBERT \cite{delobelle_2020_RobBERTDutchRoBERTabased}. However, these models do display the weaker reduction effect, and further, the absence of a significant difference between conditions for these models when presented with the full stories shows that the difference between canonical and noncanonical critical sentence stimuli is not just reduced, but disappears entirely.

It may be tempting to infer that the architecture of autoregressive transformers, and in particular, those based on the GPT-2 architecture, leads to success capturing the effect. However, it should be noted that before correction for multiple comparisons, RobBERT also successfully displays the reversal effect. In addition, not all language models had the same vocabulary, and thus, a different number of items were analyzed across models\footnote{Though it should be noted that an alternate analysis including all critical words by operationalizing the suprisal of multi-token words as the sum of their tokens' surprisals (see \citeNP{michaelov_2022_MoreHumanlikeLanguage}) shows the same qualitative results for all models except for BERTje---which performs worse.}. For these reasons, and because these models are all of various sizes and trained on several different datasets, we believe it would be premature to draw conclusions about how language model architecture impacts whether a model displays the reversal effect.

\begin{figure*}[h]
    \centering
    \includegraphics[width=0.9\textwidth]{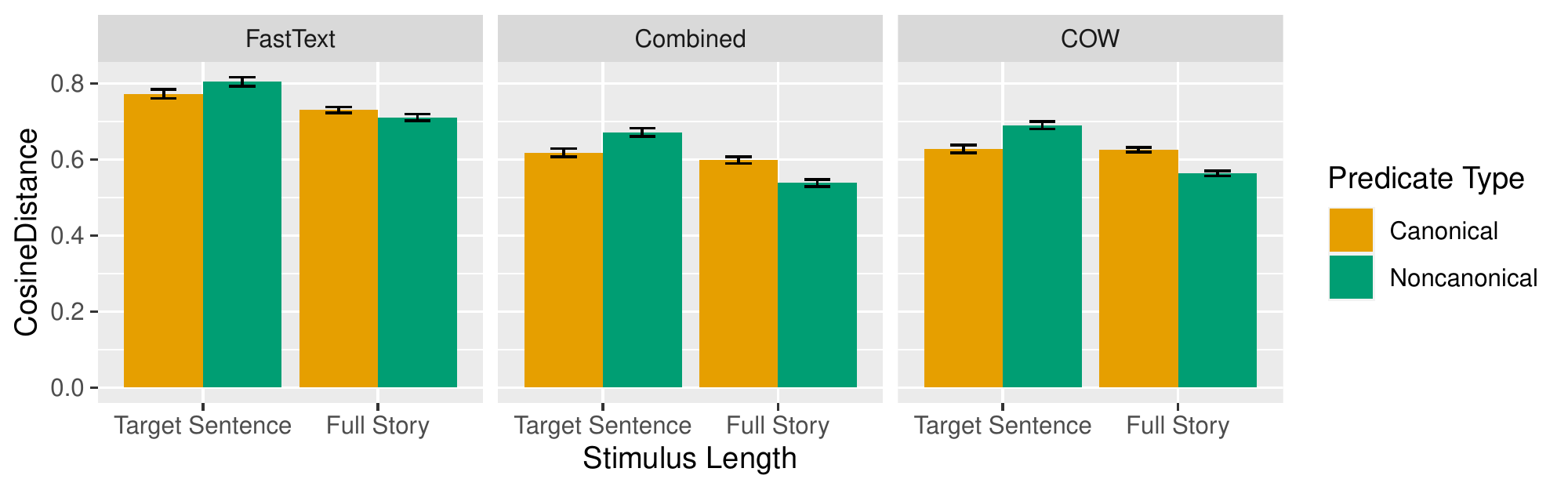}
    \caption{Cosine distance elicited by critical words for each predicate type and stimulus length.}
    \label{fig:embeddings}
\end{figure*}

\section{Experiment 2: Word Vectors}

\subsection{Cosine Distance}
In this study, we used 3 sets of pretrained word vectors: the 300-dimensional Dutch \textit{fastText} vectors \cite{grave_2018_LearningWordVectors} trained on Dutch text from Wikipedia\footnote{\url{https://nl.wikipedia.org/}} and Common Crawl\footnote{\url{https://commoncrawl.org/}} and two 320-dimensional Dutch word vectors released by \citeA{tulkens_2016_EvaluatingUnsupervisedDutch}---one trained on \textit{COW} (COrpora from the Web; \citeNP{schafer_2012_BuildingLargeCorpora}) and one trained on a \textit{Combined} corpus made up of the SoNaR corpus \cite{oostdijk_2013_Construction500MillionWordReference} and text from Wikipedia and Roularta\footnote{\url{https://www.roularta.be}}. Cosine distance was calculated (using \textit{SciPy}; \citeNP{virtanen_2020_SciPyFundamentalAlgorithms}) between the mean of the word vectors for all words in the preceding context and the word embedding for the critical word. All critical words were present in the vectors, so all experimental items were included in the analysis; it should be noted though that words in the context that were not present in the vectors were ignored when calculating cosine distance. The cosine distances for critical words in each condition are shown in Figure \ref{fig:embeddings}.

\subsection{Reduction effect}
As with language model surprisal, we constructed linear mixed-effects regressions with predicate type and stimulus length as fixed effects and story from as a random intercept. With these models, the cosine distance calculated using each set of word vectors was the dependent variable. The interaction between predicate type and stimulus length was significant for all vectors after correcting for multiple comparisons (fastText: $\chi^2(1) = 12.0, p=0.003$; Combined: $\chi^2(1) = 40.8, p<0.001$; COW: $\chi^2(1) = 66.4, p<0.001$).

\subsection{Reversal effect}
When comparing the cosine distances calculated between the embedding of the critical words and the preceding words of the critical sentence using two-tailed $t$-tests as with surprisal, there was a significant difference between canonical and noncanonical critical words for Combined and COW vectors 
(Combined: $t(116.9)=-3.48$, $p=0.004$;
COW: $t(116.5)=-4.45$, $p<0.001$), 
but not fastText vectors
(fastText: $t(118.0)={-1.96}$, $p=0.237$).

Similarly, when comparing the cosine distances between the critical word and the preceding words of the full story, there was a significant difference between canonical and noncanonical critical words for Combined and COW vectors
(Combined: $t(117.0)=4.82$, $p<0.001$;
COW: $t(117.0)=6.78$, $p<0.001$),
but not fastText vectors
(fastText: $t(117.4)={1.68}$, $p=0.418$).

\subsection{Discussion}
The cosine distances calculated from all three sets of word vectors displayed the reduction effect, and two out of three displayed the reversal effect. Thus, the results suggest that the N400 effect reported by \citeA{nieuwland_2006_WhenPeanutsFall} can be explained by lexical priming based on distributional linguistic knowledge alone.

The present study corroborates the finding of \citeA{uchida_2021_ModelOnlineTemporalSpatial}, and expands upon it in several ways. First, we explicitly tested for the reversal effect---not just whether canonical and noncanonical stimuli differ depending on whether there is a preceding story or not, but also whether the noncanonical sentence is more expected than the canonical when the story is present. Second, we found that word vector cosine distance can model the effect for multiple stimuli, not just the \textit{peanut was in love} example. Third, we found that the effect can be modeled in Dutch, the language in which the human study was carried out. And finally, we found that vectors derived from text data only (i.e., without additional information) are able to model the effect.

\section{General Discussion}

\label{ssec:human_implications}
Human comprehenders use context to update expectations about upcoming words, making a sentence that would be highly unlikely on its own more predictable than a sentence that would be relatively likely on its own. More strikingly, humans do this even when the event described is implausible, violating the constraint, for instance, that only animate, conscious entities can fall in love. The human comprehension system is quite flexible if it can update expectations about what peanuts, for example, can do, based only a story that indirectly implies the animacy of a fictional peanut.

It has often been assumed that this flexibility requires situation models that are explicitly structured \cite{venhuizen_2019_ExpectationbasedComprehensionModeling}
or involve non-linguistic world knowledge \cite{uchida_2021_ModelOnlineTemporalSpatial}. 
However, the present findings show that it is possible for purely linguistic language models model with no direct experiential grounding to update their expectations based on the linguistic context and knowledge of the statistics of language. Thus, the dynamics of event-level priming based on the distributional statistics of language may in some implicit, unspecified way approximate the effects on language comprehension previously ascribed to situation models.

In fact, the results of the present study provide evidence for an even simpler explanation. Within final sentences alone, canonical critical words were more similar to their contexts than noncanonical words, but when we include the full story context, it is the noncanonical critical words that are more similar to their contexts. It is already well-established that the amplitude of the N400 to a given word is reduced when it is semantically related to a previously-seen word \cite{bentin_1985_EventrelatedPotentialsLexical,rugg_1985_EffectsSemanticPriming,vanpetten_1988_TrackingTimeCourse,kutas_1989_ElectrophysiologicalProbeIncidental,holcomb_1988_AutomaticAttentionalProcessing,kutas_1993_CompanyOtherWords,lau_2013_DissociatingN400Effects}. Overall, then, our results show that in principle, it is possible that the pattern in the N400 responses reported by \citeA{nieuwland_2006_WhenPeanutsFall} may not rely on situation models or even event-level priming, but rather reflect some form of lexical priming. 

It may still be the case that humans use structured or semantically-rich situation models in online language comprehension (see, e.g., \citeNP{kuperberg_2020_TaleTwoPositivities}). However, the results of the study carried out by \citeA{nieuwland_2006_WhenPeanutsFall} appear to provide weaker evidence for this than previously believed. Language model predictions or even lexical priming based on language statistics appear to be sufficient to explain the effect, at least qualitatively---a valuable line of future research would be to test whether these can fully account for the effect in single-trial N400 data.

\section{Acknowledgments}
We would like to thank the anonymous reviewers for their valuable comments. This work was partially supported by a 2021-2022 Center for Academic Research and Training in Anthropogeny Annette Merle-Smith Fellowship awarded to James A. Michaelov.
 
\bibliographystyle{apacite}

\setlength{\bibleftmargin}{.125in}
\setlength{\bibindent}{-\bibleftmargin}

\bibliography{library}

\end{document}